\documentclass[conference]{IEEEtran}

\usepackage[T1]{fontenc}
\usepackage[utf8]{inputenc}
\usepackage{microtype}
\microtypesetup{protrusion=false}  

\usepackage{amsmath,amssymb,amsfonts}

\usepackage{graphicx}
\usepackage[dvipsnames]{xcolor}
\usepackage{tikz}
\usepackage{pgfplots}
\pgfplotsset{compat=1.18}

\usepackage{booktabs}
\usepackage{siunitx}
\usepackage{multirow}

\usepackage[ruled,vlined,linesnumbered]{algorithm2e}

\usepackage[colorlinks=true, citecolor=Blue, linkcolor=Blue, urlcolor=Blue]{hyperref}

\newcommand{\etal}{\textit{et al.}}

\usepackage{eso-pic}
\IEEEoverridecommandlockouts  
\begin{document}
\AddToShipoutPictureFG*{\AtPageUpperLeft{\put(\LenToUnit{0.5\paperwidth},\LenToUnit{-0.45in}){\makebox[0pt]{\parbox{0.85\paperwidth}{\centering\scriptsize
\copyright~2026 IEEE. Personal use of this material is permitted. Permission from IEEE must be obtained for all other uses, in any current or future media, including reprinting/republishing this material for advertising or promotional purposes, creating new collective works, for resale or redistribution to servers or lists, or reuse of any copyrighted component of this work in other works. Accepted to the 2026 IEEE International Conference on Robot and Human Interactive Communication (RO-MAN).}}}}}

\title{BCNav: Bearing-Conditioned Depth Policies\\for Sound Source Navigation}

\author{
  \IEEEauthorblockN{Yaozhong Kang, Jiang Wang, Takeshi Ashizawa, Benjamin Yen, and Kazuhiro Nakadai}
  \IEEEauthorblockA{Department of Systems and Control Engineering, Institute of Science Tokyo, Japan\\
    \{kangyaozhong, wangjiang, ashizawa, benjamin, nakadai\}@ra.sc.eng.isct.ac.jp}
}

\maketitle

\begin{abstract}
The ability to navigate toward sound sources extends a robot's reach beyond its visual field, enabling response to auditory events in unknown environments.
To equip robots with this capability, existing methods couple acoustic and visual information through joint audio-visual learning in acoustic simulators. However, acoustic simulation is both low-fidelity and expensive, producing a domain gap that prevents reliable real-world deployment, while the discrete action spaces inherited from grid-based simulators introduce an additional kinematic gap on physical robots.
To alleviate these issues, we propose \textbf{BCNav}, a decoupled framework that separates the acoustic module from the learned navigation policy using direction-of-arrival (DOA) estimation: an estimator provides a scalar bearing to the sound source, so the navigation policy only processes depth images and a bearing angle, two inputs whose domain gaps are well characterized.
We collect shortest-path demonstrations with calibrated bearing noise injection and train the policy via imitation learning to output continuous velocity commands directly executable on ground robots.
We demonstrate the method in simulation and on a physical robot, navigating unknown environments without any acoustic fine-tuning, prior mapping, or real-world audio data collection.
Code is available at \url{https://github.com/york1to/bcnav}.
\end{abstract}


\section{Introduction}
\label{sec:introduction}

Navigation in unknown environments is a fundamental capability of autonomous robots.
To navigate, a robot needs cues about the goal: these may be given explicitly as coordinates~\cite{anderson2018evaluation} or inferred from raw observations that hint at the goal location~\cite{sridhar2024nomad, gode2025flownav}.
In this work, we focus on scenarios where the robot must navigate in an unknown environment toward a sound source, such as a voice call or a ringing phone, using only acoustic cues.

Solving this task requires three capabilities: an acoustic module that localizes the sound source, a vision module that perceives the surrounding geometry, and a planner that translates these percepts into motion.

Existing approaches~\cite{chen2020soundspaces, younes2023catchme,chen2024sim2real} couple the acoustic and vision modules through joint learning: they train models to simultaneously learn representations of spatial acoustic signals and visual observations.
Because real-world egocentric audio-visual datasets for moving robots are scarce, these methods rely on joint audio-visual simulators~\cite{chen2020soundspaces, chen2022soundspaces2} that integrate spatial acoustic rendering with visual rendering in a single platform.
With such synthetic data, a discrete action space, and a single known sound source, this task can be solved to high success rates in simulation, either via end-to-end reinforcement learning that absorbs the planner into discrete actions~\cite{younes2023catchme}, or by learning explicit audio-visual representations paired with a separate simultaneous localization and mapping (SLAM)-based planner~\cite{chen2024sim2real}.

However, the joint audio-visual learning paradigm breaks down at deployment because spatial acoustic signals are hard to render faithfully: a sound wave undergoes emission, propagation, reflection, diffraction, and absorption before reaching the microphone, and simulating this chain requires accurate room geometry, surface materials, and wave physics.
Spatial acoustic simulators~\cite{chen2022soundspaces2} approximate these processes with geometric ray tracing, crudely simplifying or omitting material properties.
As a consequence, the audio-visual representations learned in simulation do not transfer reliably: the acoustic component either misses patterns that generalize to real environments or captures spurious simulator-specific ones.
Deploying such coupled policies end-to-end thus fails due to the acoustic domain gap, compounded by a kinematic gap between discrete simulation actions and the continuous velocity commands real robots require.
Even when the planner is replaced by a SLAM-based module, deployment still requires real-world audio recordings, a pre-built map, and a discrete action space, leaving both gaps unresolved in unknown environments.

\begin{figure}[t]
  \centering
  \includegraphics[width=0.62\columnwidth]{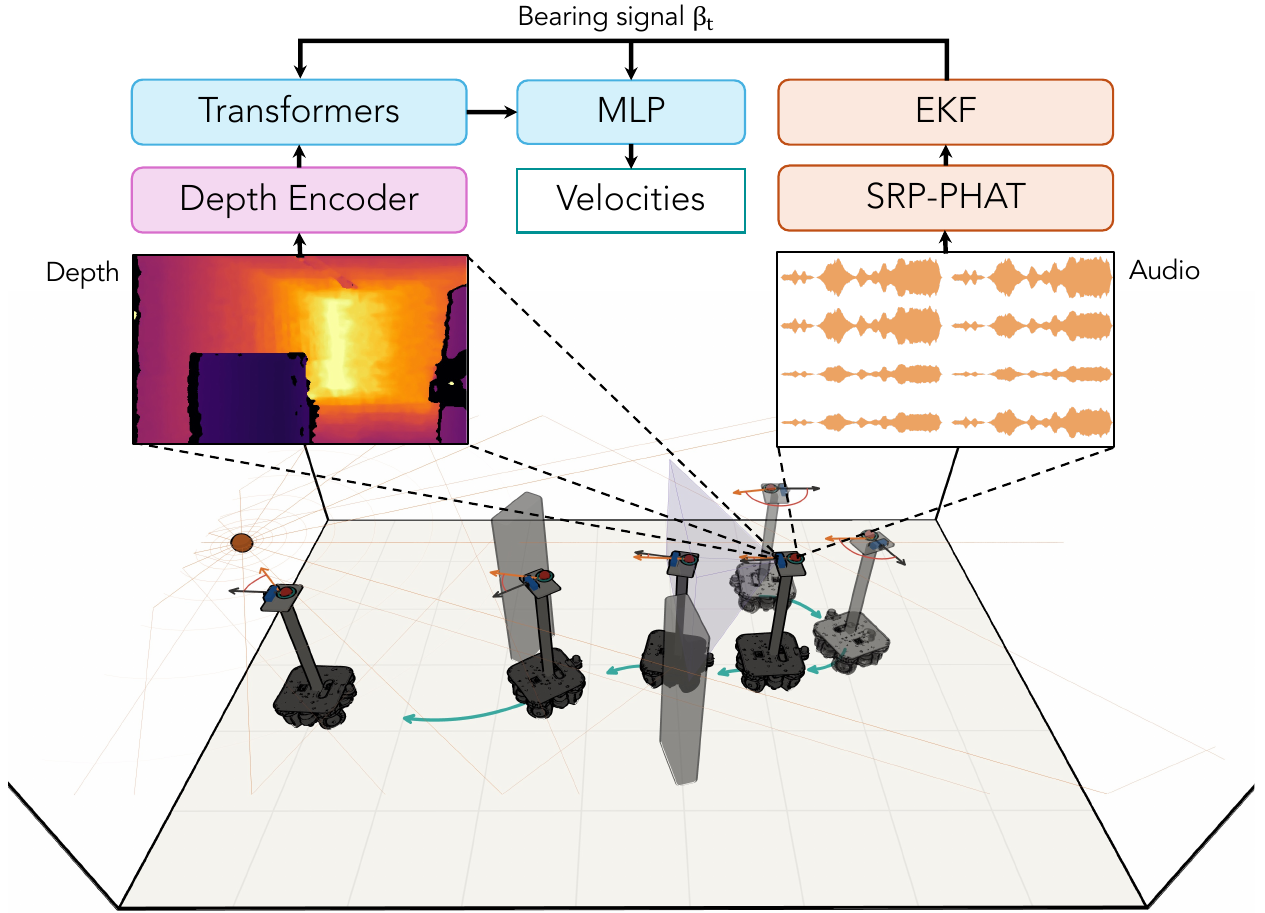}
  \caption{System overview. A classical DOA estimator provides a bearing to the sound source. The learned policy maps depth and bearing to continuous velocity commands.}
  \label{fig:overview}
\end{figure}

Given that acoustic simulation is both expensive and unreliable, we argue that the acoustic and vision modules should not be learned jointly.
Instead, we decouple the acoustic module and provide its output as a stable signal to the planner, reducing the task to standard navigation with a noisy directional cue.
The most robust and generalizable property of a sound source is its \emph{direction-of-arrival (DOA)}: both classical methods such as Steered-Response Power with Phase Transform (SRP-PHAT)~\cite{dibiase2001srpphat} and learned Sound Event Localization and Detection (SELD)~\cite{adavanne2018seldnet, shimada2021accdoa} estimate DOA reliably from commodity microphone arrays.
Reducing the audio modality to a scalar bearing removes the acoustic sim-to-real gap: the navigation policy sees only a depth image and a scalar angle, two inputs whose domain gaps are well characterized.
Our main contributions are as follows:

1) We propose a decoupled framework that places the learned/classical boundary at DOA estimation: a DOA estimator provides a scalar bearing, so the navigation policy never processes raw audio and the acoustic sim-to-real gap is sidestepped by construction, unlike joint audio-visual approaches that require retraining on real-world recordings.

2) We present a data generation pipeline paired with a continuous-control policy.
Shortest-path demonstrations with calibrated bearing noise injection are collected in simulation, and the policy is trained via imitation learning with Dataset Aggregation (DAgger) to map depth images and a scalar bearing directly to continuous velocity commands.
This approach avoids the need for lengthy online reinforcement learning while producing a policy robust to imprecise real-world DOA estimates.

3) We validate BCNav both in simulation and on a physical robot navigating in unknown environments without prior mapping or audio data collection, and characterize the system's operating envelope linking DOA accuracy to navigation success.

\section{Related Work}
\label{sec:related_work}

\subsection{Audio-Visual Navigation}

The AudioGoal task~\cite{chen2020soundspaces} trains an agent to reach a sounding object using egocentric vision and audio rendered by SoundSpaces~\cite{chen2022soundspaces2}.
The original method uses reinforcement learning with audio-visual observations and discrete actions; subsequent work extends the task to moving sources~\cite{younes2023catchme} and multi-agent pursuit, but remains within the same simulator and discrete action space.
Chen~et~al.~\cite{chen2024sim2real} present the only real-robot deployment to date.
They split the system into a learned acoustic field predictor and a classical planner, and apply frequency-adaptive filtering and noise augmentation to partially close the acoustic domain gap.
However, their learned acoustic module still requires real-world audio recordings from the target environment and a pre-built map, limiting deployment to known spaces.
In contrast, our framework eliminates the learned acoustic component altogether by reducing the audio input to a scalar bearing estimated by a classical DOA method, so the policy can be deployed in unknown environments without prior audio data or mapping.
More broadly, non-visual sensing modalities such as Wi-Fi channel state information have been combined with mobile robots for human-centric tasks like fall detection and response~\cite{chen2024falldetection}, demonstrating the value of deploying robots that react to non-visual cues in indoor environments.

\subsection{Sound Source Localization}
Classical methods such as SRP-PHAT~\cite{dibiase2001srpphat} estimate DOA from time delays across microphone pairs.
SRP-PHAT is lightweight, requires no training data, and remains robust to noise and moderate reverberation, making it suitable for real-time deployment on edge devices, but degrades under heavy reverberation and non-line-of-sight conditions.
Learning-based alternatives~\cite{adavanne2018seldnet, shimada2021accdoa} achieve strong results on real-world recordings but require labeled datasets from the target acoustic domain.
Prior audio-visual navigation work has either learned implicit audio features end-to-end~\cite{chen2020soundspaces} or predicted richer spatial representations~\cite{chen2024sim2real}, bypassing explicit DOA.
We use DOA directly as the interface between the acoustic and navigation modules, and select SRP-PHAT for deployment because it runs in real time on embedded hardware without environment-specific training data.
Because our framework treats the DOA module as a black box, it is compatible with any localization method, classical or learned.

\begin{figure*}[!t]
  \centering
  \includegraphics[width=\textwidth,height=0.15\textheight,keepaspectratio]{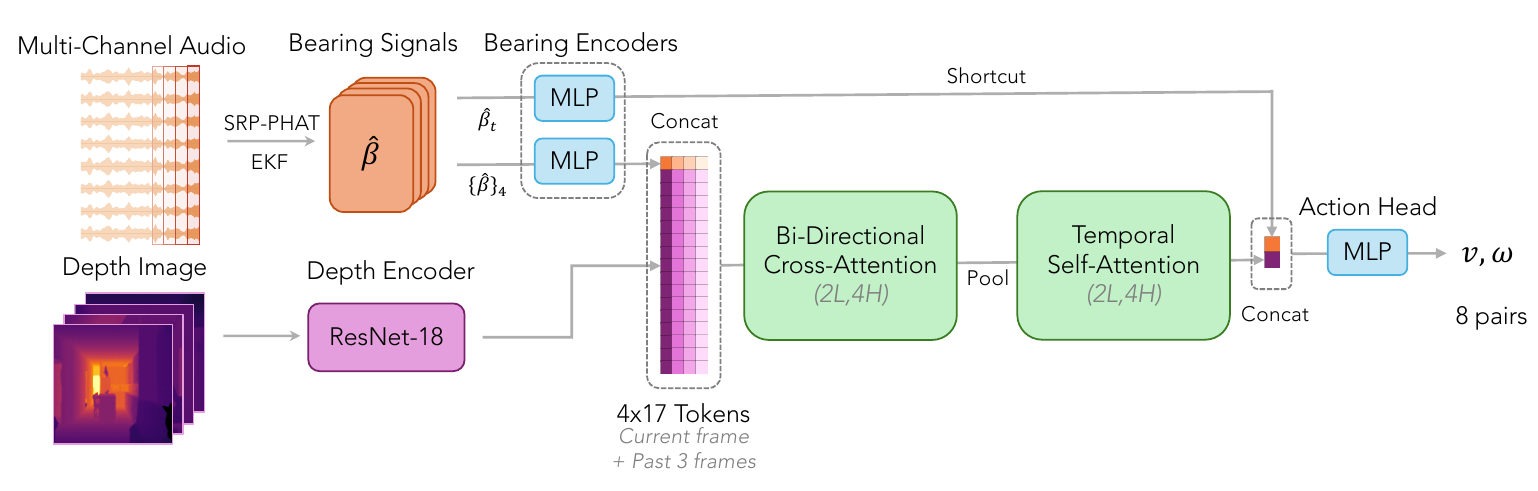}
  \caption{Policy architecture. Each timestep's depth image produces 16 spatial tokens via a DeFM ResNet-18. The bearing is encoded as a single token via sinusoidal embedding. Bidirectional cross-attention fuses the two modalities. A temporal transformer aggregates four frames. A bearing shortcut concatenated with the temporal output feeds a regression MLP that predicts 8 velocity pairs.}
  \label{fig:architecture}
\end{figure*}

\subsection{Continuous and Reactive Navigation}
Most audio-visual navigation methods inherit discrete actions from grid-based simulators~\cite{chen2022soundspaces2}.
Real robots require continuous velocity commands.
Early work on mapless navigation showed that policies can map range sensors and a goal direction directly to continuous velocities without building a map, either through reinforcement learning~\cite{tai2017virtual} or imitation learning on an expert planner~\cite{pfeiffer2017perception}.
These reactive policies implicitly learn obstacle avoidance from depth observations as a byproduct of training.
Recent methods scale this paradigm with richer representations: ViNT~\cite{shah2023vint} predicts waypoints from temporal image context, NoMaD~\cite{sridhar2024nomad} uses denoising diffusion for multimodal actions, and FlowNav~\cite{gode2025flownav} replaces diffusion with conditional flow matching for faster inference.
Sampling-based planners instead generate diverse trajectory candidates through Stein variational inference~\cite{yin2024steinmp}, optionally guided by reinforcement learning~\cite{cai2025qstac}.
However, these methods all condition on visual goals (target images) or point goals (coordinates), and none has been applied to audio-guided navigation.
Our work replaces the goal cue with an acoustic bearing and builds a reactive policy that operates directly in unknown environments, drawing on architectural insights from this line of work: a pretrained depth encoder (DeFM~\cite{patel2026defm}, self-distilled on 60M depth images), temporal context, and imitation learning.

\section{Method}
\label{sec:method}

BCNav decomposes sound source navigation into two independent modules: a classical DOA estimator that converts microphone array signals into a scalar bearing, and a learned policy that maps depth images and the bearing to continuous velocity commands.
Fig.~\ref{fig:overview} illustrates this decomposition; the navigation policy never processes raw audio.

\subsection{Problem Formulation}
\label{sec:problem}

At each decision step~$t$, the agent observes a depth image $\mathbf{d}_t \in \mathbb{R}^{1 \times 128 \times 128}$ and a bearing $\beta_t \in [-\pi, \pi]$ indicating the estimated direction to the sound source in the robot's egocentric frame.
Given a temporal context of $T{=}4$ observations $\{(\mathbf{d}_{t-k}, \beta_{t-k})\}_{k=0}^{T-1}$ (three past frames and the current frame at \SI{10}{\hertz}), the policy outputs a velocity trajectory over a horizon of $N{=}8$ steps:
\begin{equation}
  \pi_\theta\bigl(\mathbf{d}_{t- 3:t},\; \beta_{t- 3:t}\bigr) = \{(v_h, \omega_h)\}_{h=1}^{N},
\end{equation}
where $v_h$ and $\omega_h$ are the robot base's linear and angular velocities, expressed in the body frame.
The robot executes the first command and re-plans at the next step.
\subsection{Policy Architecture}
\label{sec:architecture}

Fig.~\ref{fig:architecture} shows the network architecture (17.3M parameters).
The policy encodes depth and bearing at each timestep, fuses them via cross-attention, aggregates temporal context, and predicts velocities through a regression head.

\paragraph{\textbf{Depth encoder}}
Each depth frame is converted to a log-compressed three-channel representation and processed by a DeFM ResNet-18 backbone~\cite{patel2026defm} pretrained on 60M depth images via self-distillation.
The backbone produces a $4{\times}4$ feature map, which is flattened and projected into 16 spatial tokens of dimension~256.
The backbone weights are fine-tuned during training.

\paragraph{\textbf{Bearing encoder}}
The bearing enters the network through two independent paths (Fig.~\ref{fig:architecture}).
In the \emph{token path}, each $\beta_t$ is encoded as $[\sin(\beta_t), \cos(\beta_t)]$ and projected through a multilayer perceptron (MLP, $2{\to}64{\to}256$) into a single token of dimension~256, which is concatenated with the 16 depth tokens to form 17 tokens per timestep.
In the \emph{shortcut path}, only the current $\beta_t$ is re-encoded via a separate MLP ($2{\to}256{\to}256$) and concatenated directly with the temporal context before the action head.
The shortcut preserves the bearing signal that would otherwise be diluted through cross-attention and temporal pooling.
Both paths use sinusoidal encoding to avoid the discontinuity at $\pm\pi$.

\paragraph{\textbf{Cross-modal fusion}}
At each timestep, depth and bearing tokens are fused via two-layer bidirectional cross-attention~\cite{hiller2024bixt} with four heads.
Depth tokens query the bearing for directional context, and the bearing token queries depth for geometric context.
The fused tokens are mean-pooled into a single per-timestep feature.

\paragraph{\textbf{Temporal aggregation}}
The four per-timestep features are augmented with sinusoidal positional encodings and processed by a two-layer transformer encoder with four-head self-attention, then mean-pooled into a temporal context vector.
The temporal context is concatenated with the bearing shortcut feature and projected through a linear layer ($512{\to}256$).
A three-layer MLP maps the resulting representation to the velocity trajectory $\{(v_h, \omega_h)\}_{h=1}^{N}$.

\paragraph{\textbf{Collision reflex}}
At inference, when the minimum depth in the lower field of view falls below \SI{0.2}{m}, the robot suppresses forward velocity while still executing the policy's angular velocity command.
This allows the robot to rotate away from obstacles without advancing into them.

\subsection{Expert Data and Training}
\label{sec:training}

\paragraph{\textbf{Data collection}}
We collect expert demonstrations by running a shortest-path planner with proportional-derivative (PD) tracking in SoundSpaces~2.0~\cite{chen2022soundspaces2} built on Habitat~\cite{savva2019habitat}, without rendering or collecting raw audio.
Start and goal positions are sampled using a corridor-biased strategy that favors positions along the navigable skeleton of each scene, avoiding degenerate episodes in dead-ends.
Geodesic path lengths range from \SI{2}{\meter} to \SI{15}{\meter}.
The planner provides ground-truth velocity commands at each step, which serve as the regression targets.
We collect 40K episodes with an average of 121 steps across 800 Habitat-Matterport 3D (HM3D)~\cite{ramakrishnan2021hm3d} training scenes.

\begin{figure}[t]
  \centering
  \includegraphics[width=0.70\columnwidth]{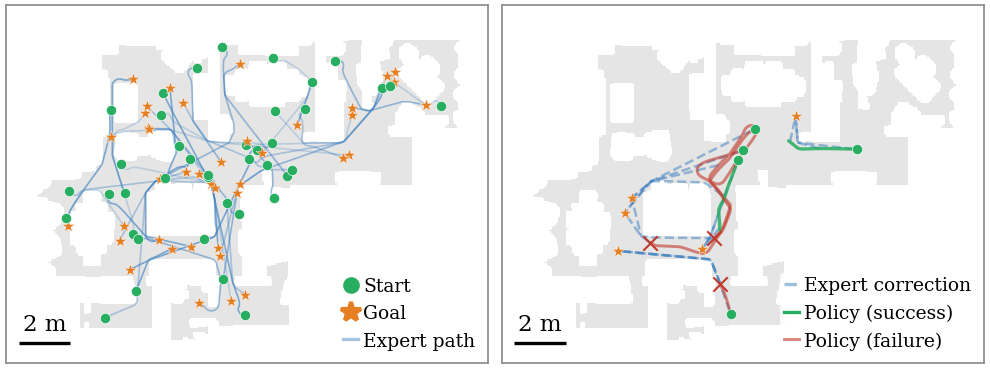}
  \caption{Training data in one HM3D scene. (a)~Expert demonstrations follow shortest paths between diverse start--goal pairs. (b)~DAgger rollouts: the policy (solid) deviates from the expert path, and expert corrections (blue dashed) are computed from the visited states back to the goal, providing on-policy training labels.}
  \label{fig:dataset}
\end{figure}

To address distributional shift, we apply DAgger~\cite{ross2011dagger}: the current policy is deployed in simulation, and expert corrections are recorded from the states the policy actually visits (Fig.~\ref{fig:dataset}).
These on-policy demonstrations teach the policy to recover from its own mistakes near walls and after wrong turns.
DAgger episodes comprise 25\% of each training batch (set empirically), mixed with the primary dataset via weighted random sampling.

\paragraph{\textbf{Bearing noise model}}
During training, the oracle bearing (the true direction to the goal) is corrupted with noise that models the empirical DOA error distribution:
\begin{equation}
  \hat{\beta}_t = \beta_t + \epsilon, \quad
  \epsilon \sim (1 - p)\,\mathcal{N}(0, \sigma^2) + p\,\mathcal{U}[-\pi, \pi],
  \label{eq:noise}
\end{equation}
where $\sigma{=}\SI{0.17}{\radian}$ (${\approx}\SI{10}{\degree}$) matches the mean absolute error (MAE) of SRP-PHAT in line-of-sight conditions, and $p{=}0.15$ models catastrophic failures under heavy reverberation, where the DOA estimate is effectively random.
This noise model bridges the gap between the deterministic oracle bearing available in simulation and the noisy DOA estimates encountered at deployment.

\paragraph{\textbf{Training objective}}
The policy is trained via behavioral cloning.
The primary loss is mean squared error on the predicted velocity trajectory:
\begin{equation}
  \mathcal{L}_\text{traj} = \frac{1}{N} \sum_{h=1}^{N} \bigl\|(v_h, \omega_h) - (v_h^*, \omega_h^*)\bigr\|^2,
  \label{eq:loss}
\end{equation}
where $(v_h^*, \omega_h^*)$ are the expert velocities, normalized to zero mean and unit variance.
An auxiliary obstacle head predicts the minimum depth in the lower field of view from the temporal context; $\mathcal{L}_\text{obs}$ is the MSE between predicted and ground-truth minimum depth. The total loss is:
\begin{equation}
  \mathcal{L} = \mathcal{L}_\text{traj} + \lambda \,\mathcal{L}_\text{obs}, \quad \lambda = 0.1.
\end{equation}
We train for 30 epochs with AdamW~\cite{loshchilov2019adamw} (learning rate $10^{-4}$, cosine schedule, batch size 256) and maintain an exponential moving average of the weights for evaluation.

\section{Simulation Experiments}
\label{sec:sim_experiments}

We evaluate the bearing-conditioned depth policy in SoundSpaces~2.0 to address three questions:
(1)~How does navigation success depend on bearing noise?
(2)~Where does SRP-PHAT DOA accuracy fall on this tolerance curve?
(3)~Does each component contribute?

\subsection{Setup}
\label{sec:sim_setup}

\paragraph{Environment.}
Evaluation uses 100 unseen HM3D validation scenes with five episodes per scene, yielding 463 valid episodes after filtering scenes with insufficient navigable area.
All experiments use random initial headings.

\paragraph{Metrics.}
We report \textbf{Success Rate} (SR), the fraction of episodes where the agent reaches within \SI{1}{m} of the goal, and \textbf{Success weighted by Path Length} (SPL)~\cite{anderson2018evaluation}.
Episodes are capped at 500 steps.

\subsection{Bearing Noise Tolerance}
\label{sec:noise_tolerance}

The central result is the operating envelope.
Under matched training/evaluation conditions (Eq.~\ref{eq:noise}, $p{=}0.15$), SR reaches 96.5\% for $\sigma \leq \SI{15}{\degree}$ and drops to ${\sim}75\%$ beyond \SI{30}{\degree}, where the bearing becomes uninformative and the policy falls back to depth-only obstacle avoidance.
This defines the system's operating envelope: any DOA method with MAE below \SI{15}{\degree} enables near-perfect navigation.

\subsection{DOA Accuracy in Simulation}
\label{sec:doa_accuracy}

We characterize SRP-PHAT accuracy on rendered audio to determine where real DOA falls on the tolerance curve.
Table~\ref{tab:doa} summarizes results from 8,000 source--receiver pairs across 16 Replica~\cite{straub2019replica} scenes (8-channel circular array, \SI{16}{kHz}).

\begin{table}[t]
  \centering
  \caption{SRP-PHAT DOA accuracy in simulation~\cite{chen2022soundspaces2}. MedAE: median absolute error.}
  \label{tab:doa}
  \sisetup{table-format=2.1}
  \begin{tabular}{@{}lSSS@{}}
    \toprule
    Condition & {MAE [\si{\degree}]} & {MedAE [\si{\degree}]} & {${<}\SI{15}{\degree}$ [\%]} \\
    \midrule
    LOS (60\%)  & 11.4 & 5.8  & 77.7 \\
    NLOS (40\%) & 37.3 & 26.4 & 36.6 \\
    \midrule
    Overall     & 21.7 & 9.0  & 61.4 \\
    \bottomrule
  \end{tabular}
\end{table}

Under acoustic line-of-sight (LOS) conditions, SRP-PHAT operates comfortably within the policy's robust range (Sec.~\ref{sec:noise_tolerance}), degrading gradually with distance but staying within the \SI{15}{\degree} threshold for typical indoor ranges.

Non-line-of-sight (NLOS) estimates are also far from random: acoustic diffraction around obstacles preserves directional information, and over a third of NLOS estimates still fall within the policy's robust range.
Degradation severity correlates with occlusion: mild occlusion (geodesic-to-Euclidean ratio 1.1--1.3) preserves usable bearings, while severe occlusion (ratio ${>}$2.0) produces effectively random estimates.

\subsection{Closed-Loop Audio Navigation}
\label{sec:closed_loop}

The preceding sections evaluate DOA accuracy and noise tolerance independently.
We now close the loop: at each step, the simulator renders room impulse responses (RIRs) via SoundSpaces~2.0, SRP-PHAT estimates the bearing from the rendered 8-channel audio, and the policy navigates using this estimated bearing instead of the oracle.

The closed-loop system achieves 71.1\% SR on HM3D validation scenes.
For context, Decentralized Distributed PPO (DD-PPO)~\cite{chen2020soundspaces} reports 82\% SR (SPL 0.63) and Chen~\etal~\cite{chen2024sim2real} report 91\% SR (SPL 0.76) on the AudioGoal task, but both evaluate with discrete actions on Replica scenes seen during training.
Our 71.1\% SR (SPL 0.709) is measured with continuous actions on unseen HM3D scenes with rendered DOA, a strictly harder evaluation setting.

The dominant failure mode is temporally correlated NLOS errors, where DOA estimates persist in the wrong direction across consecutive steps rather than averaging out as the i.i.d.\ noise model predicts.
This is consistent with the noise tolerance analysis: sustained errors above \SI{30}{\degree} push the system beyond its operating envelope.

Acoustic simulation uses reduced ray counts and omits ego-noise and material properties, producing less realistic RIRs than physical environments; SRP-PHAT, which whitens the spectrum to suppress reverberation, therefore achieves lower error on real audio than on these renderings (Sec.~\ref{sec:real_doa}).
This fidelity gap also hampers end-to-end audio-visual methods, which learn representations from the same imperfect renderings; improving NLOS robustness via learned DOA or temporal bearing filtering would raise closed-loop performance without modifying the policy.

\subsection{Ablation Studies}
\label{sec:ablations}

Table~\ref{tab:nav_results}(a) isolates the contribution of each component.
The proportional controller uses the same bearing signal but replaces the learned policy with $\omega = K_p \beta_t$, achieving only 11.8\% SR even with perfect bearing ($\sigma{=}0$).
The gap to the full system demonstrates that depth-based obstacle avoidance, not the bearing signal alone, drives navigation success.
Without the collision reflex, the learned policy frequently gets stuck against walls it cannot back away from, cutting SR by roughly a third.
Replacing direct velocity regression with waypoint prediction followed by PD tracking is even more damaging: small waypoint errors compound through the tracking controller, producing trajectories that oscillate and collide.

\begin{table}[t]
  \footnotesize
  \setlength{\tabcolsep}{3pt}
  \caption{Navigation results: (a)~simulation ablation on HM3D val and (b)~real-world trials (10 each).}
  \label{tab:nav_results}
  \begin{minipage}[t]{0.55\columnwidth}
    \centering
    {(a) Simulation ablation}\\[2pt]
    \begin{tabular}{@{}l c S[table-format=2.1] S[table-format=1.3]@{}}
      \toprule
      Config. & {$\sigma$} & {SR} & {SPL} \\
      \midrule
      Closed-loop  & {\SIrange{9}{41}{\degree}} & 71.1 & 0.709 \\
      \midrule
      Oracle       & {\SI{15}{\degree}} & 96.5 & 0.964 \\
      w/o reflex   & {\SI{15}{\degree}} & 64.2 & 0.641 \\
      w/o velocity & {\SI{15}{\degree}} & 21.3 & 0.231 \\
      P-controller & {\SI{0}{\degree}}  & 11.8 & 0.118 \\
      \bottomrule
    \end{tabular}
  \end{minipage}\hfill
  \begin{minipage}[t]{0.43\columnwidth}
    \centering
    {(b) Real-world}\\[2pt]
    \begin{tabular}{@{}l S[table-format=3.0]@{}}
      \toprule
      Setting & {SR} \\
      \midrule
      Semi-anech., static     & 100 \\
      Semi-anech., +obstacles & 70 \\
      Semi-anech., moving     & 100 \\
      \midrule
      Corridor, straight      & 100 \\
      Corridor, T-junction    & 20 \\
      \bottomrule
    \end{tabular}
  \end{minipage}
\end{table}

\section{Real-World Experiments}
\label{sec:real_experiments}

\begin{figure}[t]
  \centering
  \includegraphics[width=0.72\columnwidth]{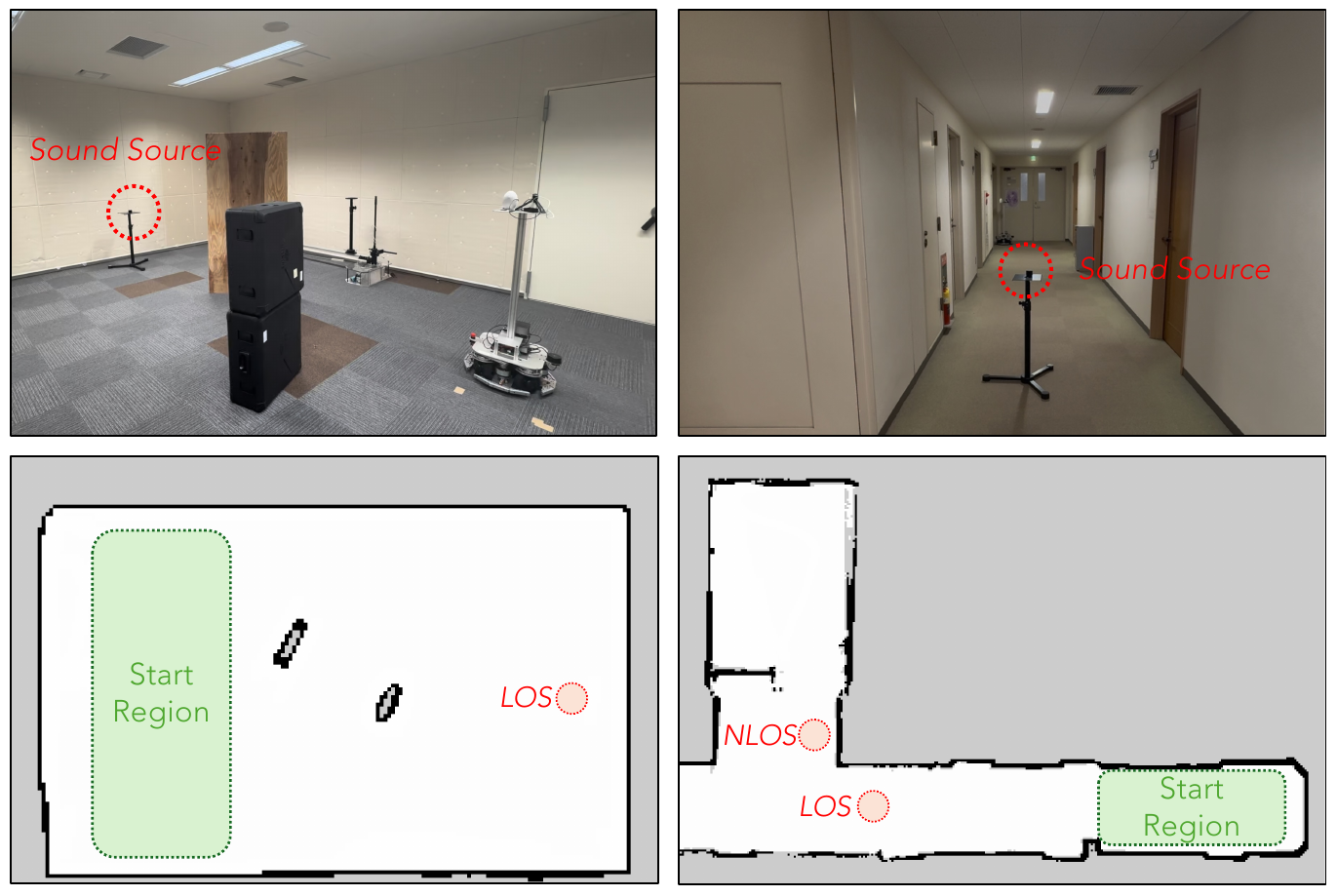}
  \caption{Real-world evaluation environments. Left: Semi-anechoic room with obstacles. Right: L-shaped corridor.}
  \label{fig:real_envs}
\end{figure}

We deploy the full pipeline on a 4WD rover equipped with an NVIDIA Jetson AGX Orin, an Orbbec Gemini~335L depth camera ($480{\times}270$\,px, \SI{10}{Hz}), and a TAMAGO-03 8-channel microphone array (\SI{36.5}{mm} radius, \SI{16}{kHz}).
We evaluate in two environments: a semi-anechoic room (\SI{6}{m}$\times$\SI{4}{m}) and an L-shaped corridor (\SI{12}{m} total length), shown in Fig.~\ref{fig:real_envs}.
The sound source plays arbitrary music through a Bluetooth speaker, with no constraints on spectral content or signal type.
Deployment requires no real-world audio data collection, white-noise calibration, or robot-noise augmentation: unlike methods that learn environment-specific audio representations~\cite{chen2024sim2real}, our system is content-agnostic, since SRP-PHAT needs only sufficient bandwidth to estimate DOA and the policy never processes raw audio.
In all trials, the sound source is deliberately placed outside the depth camera's field of view at the start of the episode.

\subsection{DOA Accuracy}
\label{sec:real_doa}

We first measure static DOA accuracy at \SI{1}{m} source distance.
In the semi-anechoic room, SRP-PHAT achieves MAE below \SI{2.4}{\degree}; in the corridor, multipath reflections raise MAE to \SI{15.2}{\degree} in the worst-case direction.
Both are within or near the policy's robust operating range (Sec.~\ref{sec:noise_tolerance}).

\begin{figure}[t]
  \centering
  \includegraphics[width=0.50\columnwidth]{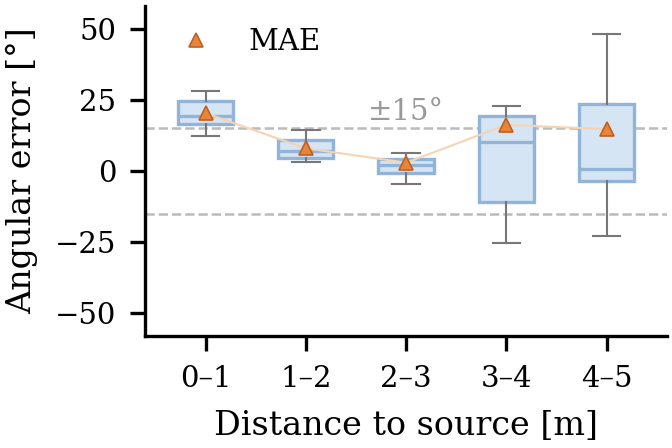}
  \caption{DOA error vs.\ source distance during real-world navigation. Dashed lines mark the $\pm\SI{15}{\degree}$ robust operating range from Sec.~\ref{sec:noise_tolerance}.}
  \label{fig:doa_error}
\end{figure}

During navigation (Fig.~\ref{fig:doa_error}), DOA error varies with distance to the source.
At 2--\SI{3}{m}, MAE drops to \SI{2.9}{\degree}.
Near-field ($<$\SI{1}{m}) and far-field ($>$\SI{4}{m}) conditions degrade accuracy to \SI{15}{\degree}--\SI{20}{\degree}, consistent with the known limitations of small-aperture arrays at extreme ranges.

\subsection{Navigation Results}
\label{sec:real_nav}

We evaluate five navigation scenarios, summarized in Table~\ref{tab:nav_results}(b).

\paragraph{Semi-anechoic room.}
In both stationary and moving source trials, the robot navigates to the sound source while avoiding obstacles.
The sim-trained policy transfers without fine-tuning: the bearing-conditioned depth policy produces smooth velocity commands at \SI{0.3}{m/s} linear and \SI{1}{rad/s} angular velocity, with heading oscillation below \SI{5}{\degree} when the source is within \SI{3}{m}.
The moving source scenario confirms that the policy reacts to bearing changes in real time rather than memorizing fixed trajectories.

The primary failure mode is collision with obstacles outside the depth camera's \SI{60}{\degree} horizontal field of view (FOV).
When the robot turns sharply in response to a large bearing update, nearby obstacles can enter the blind zone.
This is not a limitation of the learned policy but of the sensor geometry, and is mitigated by the collision reflex (Sec.~\ref{sec:architecture}) or wider-FOV depth sensors.

\paragraph{Corridor, straight path.}
With the source at the opposite end of a straight corridor segment, the robot navigates the full length (${\sim}\SI{8}{m}$) successfully.
DOA accuracy remains within the robust range throughout, consistent with the LOS condition.

\paragraph{Corridor, T-junction.}
When the source is placed around a corner in the T-junction, navigation partially succeeds.
SRP-PHAT estimates degrade severely: acoustic energy reaching the array is attenuated by the corner, and strong reverberation from the corridor walls produces spurious peaks in the spatial spectrum.
The robot approaches the sound source, but the bearing information consistently points toward the walls and changes rapidly, causing it to oscillate beside the wall.
This failure is expected from the noise tolerance analysis (Sec.~\ref{sec:noise_tolerance}): at $\sigma \geq \SI{30}{\degree}$, SR drops below 75\% even in simulation.

\paragraph{Onboard compute.}
The full pipeline runs onboard the Jetson AGX Orin (MAXN) with no offboard computation.
The policy (17.3\,M parameters, FP16) and depth-based safety filter execute in \SI{15.5}{ms} and \SI{9.2}{ms} per step (medians over 200 runs), so the navigation node replans in ${\sim}\SI{25}{ms}$, well within the \SI{10}{Hz} control budget; the SRP-PHAT front-end dominates at \SI{115}{ms} per chunk, updating the bearing asynchronously at ${\sim}\SI{9}{Hz}$, sufficient for the slowly varying bearing.
The learned policy is thus far from the compute bottleneck.

\section{Conclusion}
\label{sec:conclusion}

We presented BCNav, a modular framework for sound source navigation that decouples acoustic sensing from the learned policy via DOA estimation: reducing the audio modality to a scalar bearing lets the policy see only depth and a bearing angle, sidestepping the acoustic sim-to-real gap.
The bearing noise tolerance curve (Sec.~\ref{sec:noise_tolerance}) predicts navigation success from a DOA method's accuracy without retraining, and real-world experiments confirm zero-shot transfer to physical hardware.
Limitations remain: DOA degrades under strong NLOS, real-world validation covers only two environments, and the policy assumes a single active source; future work targets learned DOA for NLOS robustness and multi-source navigation.

\section*{Acknowledgement}
This study was carried out using the TSUBAME4.0 supercomputer at Institute of Science Tokyo.
The authors used Claude (Anthropic) for grammar checking and code editing.

\bibliographystyle{IEEEtran}
\bibliography{references}

\end{document}